\documentclass[letterpaper]{article}

\usepackage{natbib,style}  
\usepackage{url,hyperref,cleveref}
\usepackage{booktabs}
\usepackage{hyperref}
\usepackage{xcolor}
\usepackage{lipsum}
\usepackage{tabularx}
\usepackage{float}      
\usepackage{subcaption} 
\usepackage{tikz}
\usetikzlibrary{arrows.meta}
\usetikzlibrary{external}

\title{Conversable Complexity:\\ Agentic LLM Collectives as Interpretable Substrates}

\author{
    Elias Najarro$^{1}$,
    Ane Espeseth$^{2}$,
    Eleni Nisioti$^{1}$,
    Sebastian Risi$^{1,4}$,\and
    Stefano Nichele$^3$
    \\
    \mbox{}\\
    $^1$IT University of Copenhagen, Denmark \\
    $^2$University of Oslo, Norway \\
    $^3$Østfold University of Applied Sciences, Norway\\
    $^4$Sakana AI, Japan
    \\
}

\begin{document}

\makeatletter
\let\sub@origmaketitle\@maketitle
\renewcommand\@maketitle{%
  \sub@origmaketitle
  \par\vspace{1ex}%
  \begingroup\centering
  \captionsetup{type=figure}%
  \begin{minipage}{0.9\textwidth}%
  \centering
  \begin{subfigure}[b]{0.48\linewidth}
    \centering
    \includegraphics[width=0.9\linewidth]{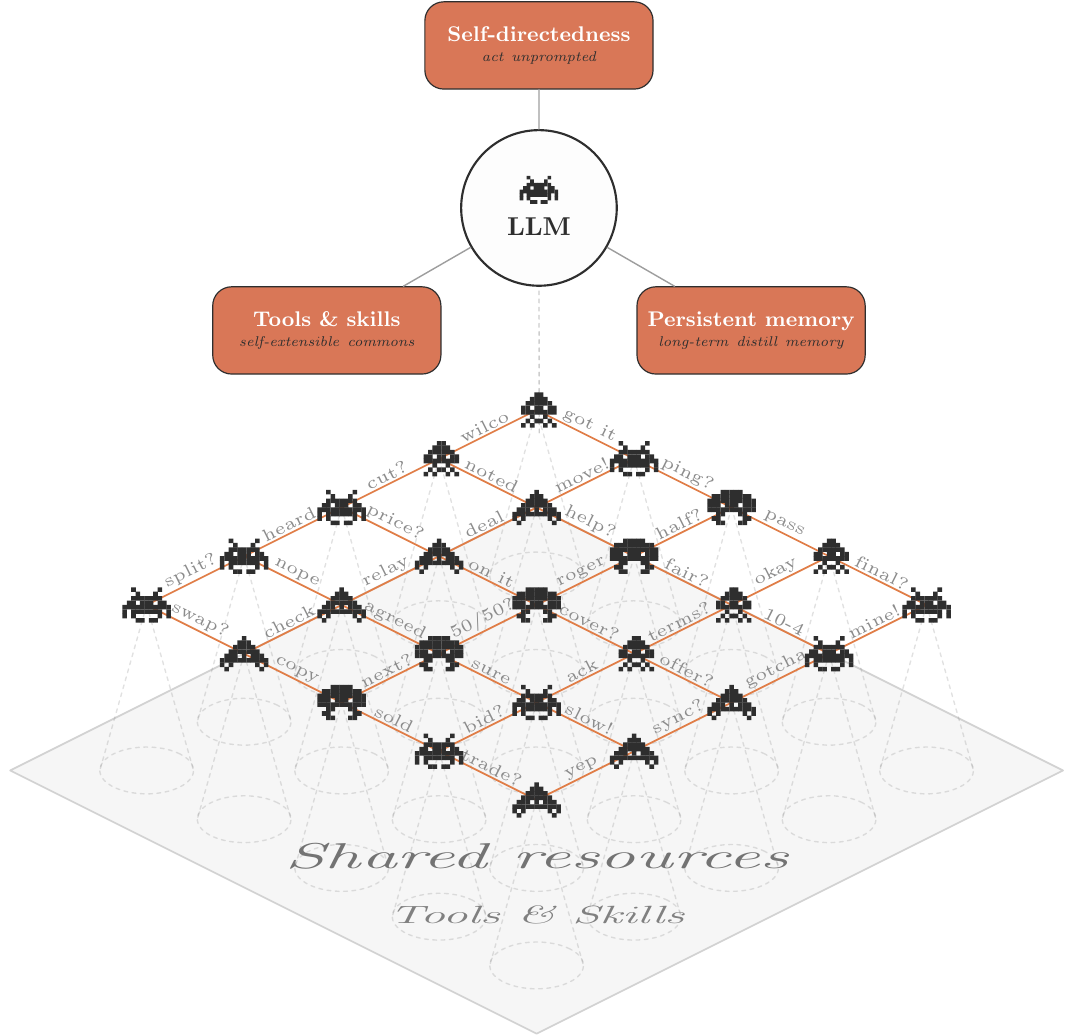}
    \label{fig:substrate-unit}
  \end{subfigure}\hspace{0\linewidth}
  \begin{subfigure}[b]{0.48\linewidth}
    \centering
    {%
      \captionsetup{width=0.9\linewidth}%
      \renewenvironment{figure}[1][]{}{}%
      \let\sthreeresizebox\resizebox
      \renewcommand{\resizebox}[3]{\sthreeresizebox{0.9\linewidth}{!}{##3}}%
      \tikzsetnextfilename{substrates_3d}%
      \tikzpicturedependsonfile{figures/substrates_3d.tex}%
      \tikzpicturedependsonfile{figures/substrates_data.tex}%
%
%

%
\definecolor{softcol}{HTML}{F13E93}     
\definecolor{hardcol}{HTML}{541A1A}     
\definecolor{wetcol}{HTML}{03AED2}      
\definecolor{agenticcol}{HTML}{2E2E2E}  

\definecolor{regdigital}{gray}{0.60}       
\definecolor{regbio}{gray}{0.52}           
\definecolor{regagentic}{gray}{0.46}       

\newcommand{\shapesoft}[2]{
  \pgfmathsetmacro{\rh}{3.0*#2}\pgfmathsetmacro{\rr}{1.95*#2}\pgfmathsetmacro{\lw}{1.0*#2}%
  \fill[white] (#1) circle (\rh pt);%
  \draw[softcol, line width=\lw pt] (#1) circle (\rr pt);%
}
\newcommand{\shapehard}[2]{
  \pgfmathsetmacro{\rh}{3.2*#2}\pgfmathsetmacro{\dd}{2.9*#2}%
  \fill[white] (#1) circle (\rh pt);%
  \filldraw[fill=hardcol, draw=black!60, line width=0.3pt]
    ([shift={(0pt,\dd pt)}]#1) -- ([shift={(\dd pt,0pt)}]#1) --
    ([shift={(0pt,-\dd pt)}]#1) -- ([shift={(-\dd pt,0pt)}]#1) -- cycle;%
}
\newcommand{\shapewet}[2]{
  \pgfmathsetmacro{\rhalo}{3.8*#2}%
  \begin{scope}[shift={(#1)}]
    \fill[white] (0,0) circle (\rhalo pt);
    \begin{scope}[x=1pt, y=1pt]
      \pgfmathsetmacro{\s}{#2}%
      \filldraw[fill=wetcol, draw=black!55, line width=0.3pt]
        (0,{3.6*\s})
        .. controls ({2.3*\s},{2.2*\s}) and ({3.0*\s},{-3.2*\s}) .. (0,{-3.6*\s})
        .. controls ({-3.0*\s},{-3.2*\s}) and ({-2.3*\s},{2.2*\s}) .. (0,{3.6*\s})
        -- cycle;%
    \end{scope}
  \end{scope}
}

\def\invpatone{0/{2,8},1/{3,7},2/{2,3,4,5,6,7,8},3/{1,2,4,5,6,8,9},%
  4/{0,1,2,3,4,5,6,7,8,9,10},5/{0,2,3,4,5,6,7,8,10},6/{0,2,8,10},7/{3,4,6,7}}
\def\invpattwo{0/{5,6},1/{4,5,6,7},2/{3,4,5,6,7,8},3/{2,3,5,6,8,9},%
  4/{2,3,4,5,6,7,8,9},5/{4,7},6/{3,5,6,8},7/{2,4,7,9}}
\def\invpatthree{0/{3,4,5,6,7},1/{1,2,3,4,5,6,7,8,9},2/{0,1,2,3,4,5,6,7,8,9,10},%
  3/{0,1,2,4,5,6,8,9,10},4/{0,1,2,3,4,5,6,7,8,9,10},5/{2,3,7,8},6/{1,2,8,9},7/{2,3,7,8}}
\def\invpatfour{0/{5},1/{4,5,6},2/{3,4,5,6,7},3/{2,3,4,5,6,7,8},%
  4/{1,2,3,5,7,8,9},5/{0,1,2,3,4,5,6,7,8,9,10},6/{0,2,8,10},7/{1,9}}
\def\invpatfive{0/{3,4,5,6,7},1/{2,3,4,5,6,7,8},2/{1,2,4,5,6,8,9},%
  3/{1,2,3,4,5,6,7,8,9},4/{0,1,2,3,4,5,6,7,8,9,10},5/{0,2,8,10},6/{2,8},7/{3,4,6,7}}
\newcommand{\drawinv}[3]{
  \begin{scope}[shift={(#1)}]
  \begin{scope}[x=1pt, y=1pt]
    \pgfmathsetmacro{\s}{0.82*#2}%
    \fill[white, rounded corners=0.6pt] ({-6*\s},{-4.6*\s}) rectangle ({6*\s},{4.6*\s});%
    \foreach \r/\cols in #3 {
      \foreach \c in \cols {
        \fill[agenticcol]
          ({(\c-5)*\s-0.52*\s},{(3.5-\r)*\s-0.52*\s})
          rectangle ({(\c-5)*\s+0.52*\s},{(3.5-\r)*\s+0.52*\s});%
      }
    }
  \end{scope}
  \end{scope}
}
\newcommand{\shapeinvA}[2]{\drawinv{#1}{#2}{\invpatone}}
\newcommand{\shapeinvB}[2]{\drawinv{#1}{#2}{\invpattwo}}
\newcommand{\shapeinvC}[2]{\drawinv{#1}{#2}{\invpatthree}}
\newcommand{\shapeinvD}[2]{\drawinv{#1}{#2}{\invpatfour}}
\newcommand{\shapeinvE}[2]{\drawinv{#1}{#2}{\invpatfive}}  

\providecommand{\lblopval}{1}
\providecommand{\regopac}{1}

\newcommand{\almk}[7]{%
  \draw[#1, draw opacity=0.35, dash pattern=on 1.1pt off 1.3pt, line width=0.35pt]
    (#3,#4,#5) -- (#3,#4,0);%
  \draw[#1, draw opacity=0.55, dash pattern=on 1.1pt off 1.3pt, line width=0.4pt]
    plot[variable=\tval, domain=0:360, samples=40]
    ({#3+0.25*cos(\tval)},{#4+0.25*sin(\tval)},0);%
  \fill[#1, opacity=0.55] (#3,#4,0) circle (0.4pt);%
  #2{#3,#4,#5}{1.0}%
  \node[#6, font=\scriptsize, text=black!80, opacity=\lblopval, inner sep=2.4pt]
    at (#3,#4,#5) {#7};%
}

\newcommand{\submark}[7]{
  \edef\subtX{#3}\edef\subtY{#4}\edef\subtZ{#5}%
  \ifx\subtX\empty\else\ifx\subtY\empty\else\ifx\subtZ\empty\else
    \pgfmathsetmacro{\subx}{0.1*\L*#3}%
    \pgfmathsetmacro{\suby}{0.1*\L*#4}%
    \pgfmathsetmacro{\subz}{0.1*\L*#5}%
    \ifdefined\animT
      \stepcounter{mkidx}%
      \pgfmathsetseed{int(1+97*\value{mkidx})}%
      \pgfmathsetmacro{\rsx}{\L*rnd}
      \pgfmathsetmacro{\rsy}{\L*rnd}
      \pgfmathsetmacro{\subx}{\rsx+\animT*(\subx-\rsx)}%
      \pgfmathsetmacro{\suby}{\rsy+\animT*(\suby-\rsy)}%
      \pgfmathsetmacro{\subz}{\animT*\subz}
    \fi
    \almk{#1}{#2}{\subx}{\suby}{\subz}{#6}{#7}%
  \fi\fi\fi
}
\newcommand{\subsoft}[5]{\submark{softcol}{\shapesoft}{#1}{#2}{#3}{#4}{#5}}
\newcommand{\subhard}[5]{\submark{hardcol}{\shapehard}{#1}{#2}{#3}{#4}{#5}}
\newcommand{\subwet}[5]{\submark{wetcol}{\shapewet}{#1}{#2}{#3}{#4}{#5}}
\newcommand{\subagentA}[5]{\submark{agenticcol}{\shapeinvA}{#1}{#2}{#3}{#4}{#5}}
\newcommand{\subagentB}[5]{\submark{agenticcol}{\shapeinvB}{#1}{#2}{#3}{#4}{#5}}
\newcommand{\subagentC}[5]{\submark{agenticcol}{\shapeinvC}{#1}{#2}{#3}{#4}{#5}}
\newcommand{\subagentD}[5]{\submark{agenticcol}{\shapeinvD}{#1}{#2}{#3}{#4}{#5}}

\newcommand{\setregion}[9]{
  \pgfmathsetmacro{\rcx}{0.1*\L*#1}\pgfmathsetmacro{\rcy}{0.1*\L*#2}%
  \pgfmathsetmacro{\rax}{0.1*\L*#3}\pgfmathsetmacro{\ray}{0.1*\L*#4}%
  \pgfmathsetmacro{\rcth}{cos(#5)}\pgfmathsetmacro{\rsth}{sin(#5)}%
  \def\rwob{#6}\def\rpha{#7}\def\rbul{#8}%
  \pgfmathsetmacro{\rbdx}{cos(#9)}\pgfmathsetmacro{\rbdy}{sin(#9)}%
}
\def\rexr{\rax*cos(\rang)*(1+\rwob*sin(3*\rang+\rpha))}
\def\reyr{\ray*sin(\rang)*(1+\rwob*sin(3*\rang+\rpha))}
\def\rxr{(\rexr)*\rcth-(\reyr)*\rsth}
\def\ryr{(\rexr)*\rsth+(\reyr)*\rcth}
\def\rdot{(\rxr)*\rbdx+(\ryr)*\rbdy}
\newcommand{\rblob}[1]{%
  plot[variable=\rang, domain=0:360, samples=140]
    ({\rcx + #1*( (\rxr) + \rbul*max(0,\rdot)*\rbdx )},
     {\rcy + #1*( (\ryr) + \rbul*max(0,\rdot)*\rbdy )},
     0)%
}
\newcommand{\planelabel}[4]{%
  \begin{scope}[x={(1cm,0cm)}, y={(0cm,1cm)}, z={(0cm,0cm)},
                cm={\projAx,\projAy,\projBx,\projBy,(0cm,0cm)}]
    \node[transform shape, #3] at ({0.1*\L*#1},{0.1*\L*#2}) {#4};
  \end{scope}%
}

\newcommand\drawregionoverlay{%
    \ifdefined\fastpreview\else
    \begin{scope}
      \clip (0,0,0) -- (\L,0,0) -- (\L,\L,0) -- (0,\L,0) -- cycle;
      \ifdefined\animT \begin{scope}[opacity=\regopac, transparency group]\fi

      \setregion{3.6}{6.7}{4.4}{2.6}{8}{0.12}{15}{0.55}{132}
      \fill[regdigital, opacity=0.045]                   \rblob{1.00} -- cycle;
      \draw[regdigital, opacity=0.40, line width=0.55pt] \rblob{1.00} -- cycle;
      \draw[regdigital, opacity=0.25, line width=0.48pt] \rblob{0.70} -- cycle;
      \draw[regdigital, opacity=0.15, line width=0.42pt] \rblob{0.42} -- cycle;

      \setregion{4.8}{1.4}{6.2}{2.2}{3}{0.11}{60}{0.50}{262}
      \fill[regbio, opacity=0.045]                       \rblob{1.00} -- cycle;
      \draw[regbio, opacity=0.40, line width=0.55pt]     \rblob{1.00} -- cycle;
      \draw[regbio, opacity=0.25, line width=0.48pt]     \rblob{0.70} -- cycle;
      \draw[regbio, opacity=0.15, line width=0.42pt]     \rblob{0.42} -- cycle;

      \setregion{9.05}{8.35}{1.7}{2.1}{-30}{0.08}{130}{0.22}{50}
      \fill[regagentic, opacity=0.045]                    \rblob{1.00} -- cycle;
      \draw[regagentic, opacity=0.40, line width=0.55pt] \rblob{1.00} -- cycle;
      \draw[regagentic, opacity=0.25, line width=0.48pt] \rblob{0.70} -- cycle;
      \draw[regagentic, opacity=0.15, line width=0.42pt] \rblob{0.42} -- cycle;
      \ifdefined\animT \end{scope}\fi
    \end{scope}
    \fi
}

\begin{figure}[t!]
  \def\yaw{30}%
  \pgfmathsetmacro{\projAx}{0.933*cos(\yaw)}
  \pgfmathsetmacro{\projAy}{0.933*0.364*sin(\yaw)}%
  \pgfmathsetmacro{\projBx}{-0.933*sin(\yaw)}
  \pgfmathsetmacro{\projBy}{0.933*0.364*cos(\yaw)}%
  \centering
  \resizebox{\columnwidth}{!}{%
  \begin{tikzpicture}[
      x={(\projAx cm,\projAy cm)},   
      y={(\projBx cm,\projBy cm)},   
      z={(0cm,0.70cm)},              
      >={Latex[length=2.0mm, width=1.6mm]},
      line cap=round, line join=round,
    ]

    \def\L{6}        
    \def\E{0.6}      

    \ifdefined\animT
      \clip ([xshift=-3.7cm, yshift=-1.6cm]0,0,0)
        rectangle ([xshift=6.3cm, yshift=7.4cm]0,0,0);
    \fi

    \draw[gray!22, line width=0.5pt] (\L,0,0) -- (\L,\L,0) -- (0,\L,0);
    \draw[gray!22, line width=0.5pt] (0,0,0) -- (0,0,\L);
    \draw[gray!22, line width=0.5pt] (\L,0,0) -- (\L,0,\L);
    \draw[gray!16, line width=0.5pt] (\L,\L,0) -- (\L,\L,\L);
    \draw[gray!18, line width=0.5pt]
      (0,0,\L) -- (\L,0,\L) -- (\L,\L,\L) -- (0,\L,\L) -- cycle;
    \drawregionoverlay

    \planelabel{4.3}{1.6}{anchor=center, font=\large\itshape, text=black!55}{Biological Complexity}
    \planelabel{3.2}{7.4}{anchor=center, font=\large\itshape, text=black!55}{Good Old-Fashioned}
    \planelabel{3.2}{6.2}{anchor=center, font=\large\itshape, text=black!55}{Artificial Life}
    \planelabel{9.2}{8.6}{anchor=center, font=\large\itshape, text=black!55}{Complex}
    \planelabel{9.2}{7.1}{anchor=center, font=\large\itshape, text=black!55}{and interpretable}

    \draw[->, black!70, line width=0.7pt] (0,0,0) -- (\L+\E,0,0)
      node[pos=0.52, sloped, rotate=-2, anchor=north, yshift=-6pt,
           font=\footnotesize\itshape, text=black!95] {Basic Unit Complexity}
      node[pos=1, sloped, anchor=north, font=\tiny, text=black!70] {high};   
    \draw[->, black!70, line width=0.7pt] (0,0,0) -- (0,\L+\E,0)
      node[pos=0.52, sloped, rotate=-1, anchor=north, yshift=-6pt,
           font=\footnotesize\itshape, text=black!95] {Emergent Complexity}
      node[pos=1, sloped, anchor=north, font=\tiny, text=black!70] {high};   
    \draw[->, black!70, line width=0.7pt] (0,\L,0) -- (0,\L,\L+\E);            

    \node[font=\tiny, text=black!70, anchor=north, yshift=-2pt] at (0,0,0) {low};
    \node[font=\tiny, text=black!70, anchor=south east, yshift=-6pt] at (0,\L,\L+\E) {high}; 

    \input{figures/substrates_data}

    \begin{scope}[x={(1cm,0cm)}, y={(0cm,1cm)}, z={(0cm,0cm)}]
      \node[font=\footnotesize\itshape, text=black!95, rotate=90]
        at (-3.4, 3.9) {Interpretability};
    \end{scope}

    \begin{scope}[x={(1cm,0cm)}, y={(0cm,1cm)}, z={(0cm,0cm)}]
      \def\ls{1.10}
      \pgfmathsetmacro{\gA}{0.85*\ls}
      \pgfmathsetmacro{\gB}{0.73*\ls}
      \pgfmathsetmacro{\ld}{0.42*\ls}
      \pgfmathsetmacro{\cw}{1.05*\ls}
      \pgfmathsetmacro{\toA}{0.16*\ls}
      \pgfmathsetmacro{\toB}{0.22*\ls}
      \pgfmathsetmacro{\axtipx}{(\L+\E)*\projAx}
      \def\agw{0.62}
      \def\rpad{0.28}
      \pgfmathsetmacro{\bxr}{\axtipx}
      \pgfmathsetmacro{\lx}{\bxr-\rpad-\agw-\toB-\cw}%
      \def\ly{-0.55}
      \pgfmathsetmacro{\bxl}{\lx-0.20}
      \pgfmathsetmacro{\byt}{\ly+0.22}
      \pgfmathsetmacro{\byb}{\ly-\ld-0.22}
      \draw[black!45, line width=0.35pt, rounded corners=2pt]
        (\bxl,\byb) rectangle (\bxr,\byt);
      \shapesoft{\lx,\ly}{\gA}
        \node[anchor=west, font=\scriptsize, text=black!80, inner sep=0pt]
          at (\lx+\toA,\ly) {Soft};
      \shapehard{\lx+\cw,\ly}{\gA}
        \node[anchor=west, font=\scriptsize, text=black!80, inner sep=0pt]
          at (\lx+\cw+\toA,\ly) {Hard};
      \shapewet{\lx,\ly-\ld}{\gA}
        \node[anchor=west, font=\scriptsize, text=black!80, inner sep=0pt]
          at (\lx+\toA,\ly-\ld) {Wet};
      \shapeinvE{\lx+\cw,\ly-\ld}{\gB}
        \node[anchor=west, font=\scriptsize, text=black!80, inner sep=0pt]
          at (\lx+\cw+\toB,\ly-\ld) {Agentic};
    \end{scope}

  \end{tikzpicture}%
  }
  \label{fig:substrates-3d}
\end{figure}
    }%
  \end{subfigure}\par
  \vspace{0.5ex}%
  \setcounter{figure}{0}%
  \captionsetup{width=0.96\linewidth}%
  
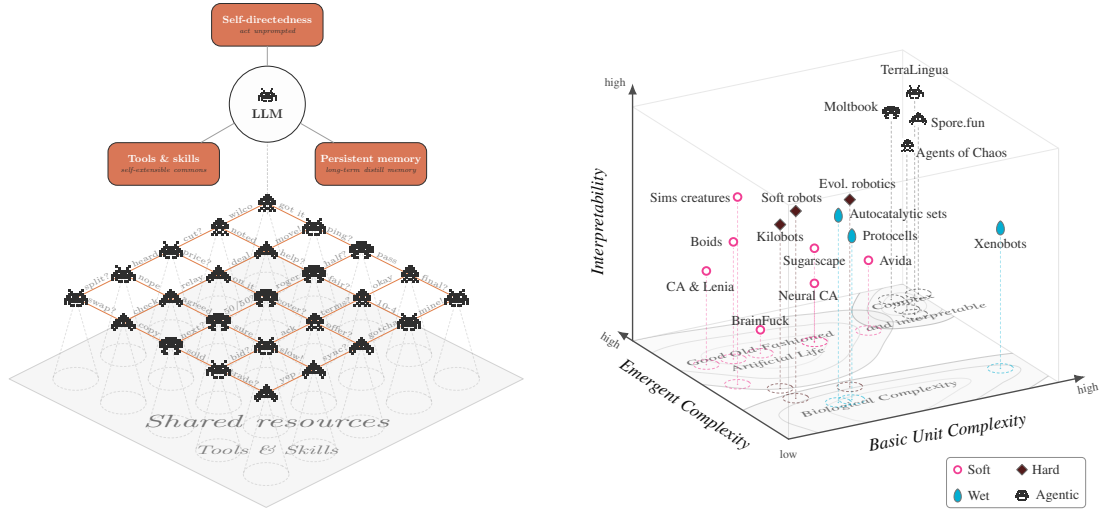
\captionof{figure}{\textbf{Agentic LLM collectives as a new substrate for Artificial
  Life.} ALife has long studied life-like phenomena in computational, physical,
  and biochemical substrates---\textit{soft}, \textit{hard}, and \textit{wet}. We argue for a fourth class: collectives of
  \emph{agentic} LLMs, language models endowed with persistent memory, access to a shared self-extensible pool of tools and skills, and the capacity to act unprompted \textit{(left)}. Their units are individually complex, yet because the agents interact in natural language the collective remains open to inspection. This places the agentic substrate in a region few substrates reach: complex yet interpretable \textit{(right)}.}%
  \label{fig:substrate}%
  \end{minipage}%
  \par\endgroup
  \vspace*{9mm}%
  \noindent\makebox[\dimexpr(\textwidth-\columnsep)/2\relax]{\Large\bfseries Abstract}\par
  \vspace{3mm}%
}
\makeatother
\maketitle

\begingroup\baselineskip=\dimexpr\baselineskip\relax plus.25\baselineskip\relax
\noindent Complexity and interpretability rarely coincide: systems rich enough for complex behaviours to emerge are usually too opaque to question, while transparent ones are too simple for anything complex to emerge. A single large language model (LLM) is a static artefact, hardly exhibiting any of the emergent properties we associate with life. This changes through interaction: populations of LLMs display emergent dynamics absent from isolated models. Furthermore, LLMs can be endowed with persistent memory, tools and shared skills, and the capacity to initiate actions unprompted, i.e., turning LLMs \emph{agentic}. In this paper, we argue that such collectives of agents can serve as a computational substrate for Artificial Life (ALife) research. Critically, since the agents communicate in natural language, their collective behaviour can be directly interrogated by examining textual traces and asking the agents themselves. We outline the notion of interpretability in language-model research and extend it for collectives of agents. Lastly, we survey recent examples of agentic LLM collectives that already instantiate the idea of \textit{agentic substrates}, from controlled experiments to deployments in the wild.\par\endgroup

\section{Introduction}
\vspace{1mm}

LLM collectives have become a popular modelling approach in computational social science (CSS), where they are widely used to investigate social phenomena such as opinion dynamics and social polarisation \citep{Grossmann2023Jun, Ashery2025May}. 
The modelling approach typically starts from a dataset that holds the social dynamics that the LLM collective is meant to reproduce or statistically account for.

In this regard, CSS uses LLM collectives in a top-down manner to explain \emph{existing} social phenomena, much as biology is top-down in seeking to understand \emph{existing} organisms.
On the other hand, Artificial Life (ALife) typically follows a bottom-up epistemic approach. Rather than accounting for one particular system, ALife asks which properties of a computational substrate make it life-like: which properties give rise, in a bottom-up manner, to the emergence of structural and functional complexity \cite{bedau2003artificial}. Its value is explanatory: a market bubble, a migration wave, or a shared convention can be investigated by testing whether simple local interactions suffice to grow it, and then perturbing those interactions to isolate which conditions drive it. In this paper we pursue this bottom-up perspective, studying collectives of LLM-based agents as a substrate for ALife research.

We begin from the premise that LLMs, taken in isolation, are not inherently life-like. An isolated LLM is an inert artefact: a complex but largely static model produced through top-down engineering. A central argument of this paper is that the situation changes when many such models interact. Collectives of LLM-based agents can exhibit dynamics that are central to ALife \citep{nisioti2024text, Nisioti2024Jul}, including adaptation, coordination, norm formation, and cultural transmission. In this view, the life-like properties emerge not from the individual model, but from the interactions among agents when placed in a shared environment. 

A useful analogy is that of biological cells in multicellular organisms. Each cell contains enormous internal complexity, yet at the level of the organism this complexity is abstracted away. Cells become components in a larger collective whose behaviour depends primarily on patterns of interaction. LLM agents can be viewed similarly. While their internal mechanisms remain complex and opaque, they can be treated as interacting units within a higher-level dynamical system. The focus therefore shifts from the architecture of a single model to the collective processes that emerge between models.


In the spirit of classical artificial societies \citep{Epstein1996Oct}, LLMs have been used to study diverse aspects of collective behaviour \citep{burton2024large}. By contrast to that line of work, we emphasise here the relevance of the \emph{agentic} aspect of the substrate's constitutive units. We use ``agentic'' to denote a computational unit in which an LLM is endowed with persistent memory, tool use, and self-directedness. Persistent memory and tools let the unit \emph{externalise cognition}---offloading cognitive mechanisms onto external structure and providing a substrate for culture---while self-directedness lets it act beyond purely reactive behaviour, for instance choosing on its own to take an action or to explore in some direction without being explicitly prompted.

In this perspective paper, we treat the agentic LLM as the basic computational unit of a substrate. We argue that collectives of such units provide an experimental testbed for bottom-up exploration and hypothesis testing of which properties---such as autonomy, self-modification, cognitive architecture, symbiogenesis, and resource constraints---are important for life-like behaviour to emerge.
Critically, because interaction is mediated through natural language, this substrate is also interpretable, making collective dynamics directly accessible to observation.

\paragraph{What constitutes a good testbed?}  
We propose that a substrate has high potential as an experimental testbed for ALife research to the extent that it satisfies five properties drawn from three traditions. In the philosophy of model organisms \citep{ankeny2020model, leonelli2013}, we inherit the requirement that the substrate has \textbf{(i)} a clearly stated representational target and scope: which phenomena the substrate is meant to illuminate, and to which other systems the results are expected to transfer; and \textbf{(ii)} standardisation sufficient for cross-laboratory comparability across runs. From the philosophy of computer simulation \citep{winsberg2010science, weisberg2013simulation}, we inherit \textbf{(iii)} the requirement that emergent results are distinguishable from artefacts: we can only make claims about a phenomenon observed in the substrate if it is robust to variation of incidental implementation choices. From Artificial Life itself \citep{langton1989artificial, bedau2000open}, we inherit \textbf{(iv)} micro-level interpretability, so that observed macro-phenomena can be traced back to manipulable micro-conditions; and  \textbf{(v)} support for generative evaluation---exploring ``life as it could be''---rather than only descriptive evaluation of existing systems.

\section{Artificial Life Substrates}
\vspace{1mm}

Artificial Life is practised across many substrates, each suited to testing different conjectures about life. We first review the existing \emph{soft}-ware, \emph{hard}-ware, and \emph{wet}-ware substrates, then propose a fourth: the \emph{agentic substrate}, built from collectives of interacting LLM agents. Such collectives run in software and are in that sense a \textit{soft} substrate, though through their tools they can also naturally interface with the physical world; we propose that this ability, together with their distinctive coincidence of high complexity and high interpretability, makes it useful to cluster instances of this substrate as a class of their own. In each case a substrate's affordances determine which conjectures it can test, a mapping we organise in Table~\ref{tab:conjectures}.

\vspace{0.5em}
\subsection{Existing Substrates: Soft, Hard \& Wet}

Artificial Life research draws on a range of substrates---many developed before ALife was a field, and only some devised specifically for it. A standard taxonomy divides them into \emph{soft}, \emph{hard}, and \emph{wet}: realised in software, in hardware, and in biochemistry, respectively \citep{bedau2007artificial}.

\paragraph{Soft Substrates.} The majority of substrates run entirely in software. \emph{Cellular automata} place a simple unit---a cell holding a discrete or, in continuous variants such as Lenia and Flow Lenia \citep{chan2019lenia, plantec2023flowlenia}, a real-valued state---on a grid and update it from its local neighbourhood, with \textit{neural} cellular automata replacing the hand-specified rule by a learned one. \emph{Agent-based models}, or ABMs, such as Sugarscape \citep{Epstein1996Oct}, Schelling's segregation model \citep{Schelling1971DynamicMO}, and Boids \citep{reynolds1987flocks} consist of units, or agents, that follow fixed, hand-specified rules while interacting in a shared spatial environment---in contrast to modern RL and LLM agents whose behaviour is driven by a learned policy rather than a hand-crafted update rule. \emph{Digital evolution} systems such as Tierra and Avida \citep{ray1991tierra, ofria2004avida}, foreshadowed by the program-combat game Core War \citep{Dewdney1984Feb}, take the unit to be a self-replicating program whose genome of machine instructions is copied, with mutation, to its offspring under selection, while \emph{evolved virtual creatures} such as Sims' creatures \citep{sims1994evolved} co-evolve a body and its controller inside a physics simulation. \emph{Multi-agent reinforcement learning}, including large-scale worlds such as Neural MMO \citep{suarez2019neural}, instead gives each agent a policy that adapts through reward-driven learning. Software substrates are used to simulate self-organisation and morphogenesis, evolutionary dynamics and the growth of complexity, division of labour, emergent communication, and ecological interaction among populations \cite{bedau2003artificial}.

\paragraph{Hard Substrates.} Other substrates give their units a physical body. \emph{Evolutionary and swarm robotics} \citep{lipson2000automatic, rubenstein2014programmable} evolve or coordinate morphologies and controllers that must act under real forces, contact, sensing noise, and energy limits, and physical \emph{soft robots} \citep{rus2015design} extend this to deformable bodies. Because their environment is the world rather than a model of it, hard substrates are used to study embodiment, the coupling of morphology to control, sensorimotor adaptation, and collective behaviour in physical swarms.

\paragraph{Wet Substrates.} A third class is built from molecules. Research on the origin of life and on synthetic biology uses \emph{protocells}, \emph{autocatalytic sets}, and chemotactic droplets \citep{egbert2010metabolism}, taking the unit to be a molecular assembly whose dynamics follow reaction kinetics, self-assembly, and thermodynamics, with heredity---where present---carried by template molecules or by compositional inheritance. Closer to multicellular life, \emph{xenobots} are motile machines assembled from living frog cells that self-organise and locomote \citep{blackiston2021cellular}. Wet substrates are used to simulate self-assembly, metabolism, self-maintenance, and the transition from non-living to living matter.

\begin{table*}[t!]
  \centering
  \footnotesize
  \setlength{\tabcolsep}{4pt}
  \renewcommand{\arraystretch}{1.0}
  \begin{tabularx}{\linewidth}{@{} p{0.18\linewidth} X X @{}}
    \toprule
    \textbf{Conjecture type} & \textbf{Previous investigations in ALife substrates} & \textbf{New avenues for research with  \textit{Agentic} Substrates} \\
    \midrule
 
    \textbf{Individuation} \newline {\scriptsize\itshape Autonomy; self-organisation}
      & \textit{Computational autopoiesis} asks whether an individual can be constituted purely by its own self-producing organisation: the Protobe lattice model \citep{varela1974autopoiesis} and its CA reconstructions \citep{mcmullin1997rediscovering} realised minimal self-maintaining boundaries.
      
    \textit{Chemotactic protocells} tied self-maintenance to behaviour \citep{egbert2010metabolism}.
      & The candidate individual can be distributed across model weights, persistent memory, and tool affordances. Self-maintenance (if it occurs) can occur through the agent's own linguistic actions on these components.  \\
    \addlinespace[1pt]
 
    \textbf{Inheritance} \newline {\scriptsize\itshape Adaptation (evolution)}
      & Tierra \citep{ray1991tierra} and Avida \citep{ofria2004avida} are digital-evolution systems in which a self-replicating program is copied, with mutation, under selection; they have been used to study how evolutionary dynamics such as parasitism, arms races, and the growth of complexity can arise from a heritable digital medium alone. & The agents carry rich digital media --- system prompts, memory files, and configurations, that can be passed to descendants. Self- or other-modification of system prompts and context files can act as mutation operations, and pretraining acts as a shared baseline across generations. \\
    \addlinespace[1pt]
 
    \textbf{Externalised cognition} \newline {\scriptsize\itshape Behaviour; information}
      & Stigmergy \citep{theraulaz1999brief} uses traces left in a shared environment as a coordination memory; niche construction \citep{odling1996niche} extends this to organisms reshaping their own selective environment, so that modifications are inherited ecologically rather than genetically. Morphological computation offloads control onto the body itself, as in passive dynamic walkers \citep{mcgeer1990passive}, while Physarum and reaction--diffusion computing \citep{adamatzky2010physarum} offload it onto the dynamics of a physical medium.
      & Agents have access to cognitive scaffolds in natural language --- persistent memory files, scratchpads, tool outputs --- which can be updated and shared between units. Because these media are readable and editable by the experimenter, their impact on the collective's behaviour can be tested with interventions and ablation replays. \\
    \addlinespace[1pt]
 
    \textbf{Role differentiation} \newline {\scriptsize\itshape Adaptation (development); artificial societies}
      & Several substrates have been used to study differentiation arising from initially uniform units: Lenia and Flow Lenia \citep{chan2019lenia, plantec2023flowlenia} grow differentiated morphology from a single homogeneous update rule; and Sugarscape \citep{Epstein1996Oct} produces social stratification from agents following identical rules.
      & Roles can be set through natural-language system prompts and modified through interaction with agents and the environment. Pretraining again acts as a shared baseline, and differentiation is observable directly in system prompts, and indirectly in individual agent behaviours. \\
    \addlinespace[1pt]
 
    \textbf{Multi-scale} \newline {\scriptsize\itshape Self-organisation; information; artificial societies}
      & The Avida multicellularity experiments \citep{goldsby2014evolutionary} model an evolutionary transition in individuality: once selection acts on groups, formerly independent replicators become differentiated components of a higher-level unit, so that a new organisational scale emerges within the digital substrate.
      & Inter-agent communication happens in natural language. Through repeated interactions, higher-level structures (norms, coalitions, hierarchies) can take form in the interaction record and through collaborative artifacts (e.g. shared agreement documents). These may or may not display downward causation, constraining the individuals' behaviours in new ways. The experimenter is also able to explore tunable parameters such as bandwidth of communication between agents. \\
    \addlinespace[1pt]
 
    \textbf{Symbol grounding} \newline {\scriptsize\itshape Information; behaviour}
      & A line of work has asked how arbitrary signals come to carry shared meaning: the Talking Heads experiment \citep{steels1999talking} grounded an emergent lexicon in situated interaction games; models of evolutionary language games \citep{cangelosi2002simulating} and iterated learning \citep{kirby2008cumulative} show compositional structure accumulating through cultural transmission.
      & Depending on the LLM model, symbols can start out grounded only in the text distribution, or in multi-modal embeddings. With situated use, coupling symbols to tools, actions, and consequences in a shared environment, these symbols could be re-grounded. The agentic collective may also coin new conventions based on their coordination history. All these changes in symbol meaning can be inferred directly from the dialogue between agents.\\
    \addlinespace[1pt]
 
    \textbf{Open-endedness} \newline {\scriptsize\itshape Adaptation (evolution); information; living technology}
      & Open-endedness has been studied in substrates designed to sustain novelty, such as PolyWorld \citep{yaeger1994polyworld}, Geb \citep{channon2001geb}, Chromaria \citep{soros2014chromaria}, and Picbreeder \citep{secretan2011picbreeder}. 
      & Recombination happens over the open vocabulary of natural language, with the creation of new roles, tools and artifacts. Selection pressure can be drawn from open environments (markets, users, other agents) rather than predefined fitness. \\
 
    \bottomrule
  \end{tabularx}
  \caption{Seven areas of conjectures central to ALife research. It highlights some examples of how existing substrates have engaged with them, and outlines how Agentic Substrates can broaden the scope of investigation. Each conjecture type is thematically linked to several areas of ALife research \citep{aguilar2014past}, here highlighted in italics.}
  \label{tab:conjectures}
\end{table*}

\subsection{Agentic Substrates}
\label{sec:agentic-ecologies}

We define an \emph{agentic substrate} as a population of \emph{agentic units}---pretrained language models, each endowed with \textbf{(i)} a persistent memory that distils salient information from its context into a structured store outlasting any single context window; \textbf{(ii)} access to a shared, self-extensible commons of tools and skills; and \textbf{(iii)} the self-directedness to act unprompted, or to decline a trigger. The population is in turn \textbf{(iv)} coupled through a connectivity structure that fixes a notion of locality and the channels (natural language) along which they communicate; and \textbf{(v)} embedded in a shared, persistent environment.

When agentic units (iii) are provided with a shared pool of tools and skills (ii), \emph{self-extensible commons} may arise: units author tools and skills and deposit them where others inherit and recombine them, so that the space of available affordances grows endogenously rather than being fixed in advance. Open-endedness, long sought but rarely instantiated \citep{bedau2000open, taylor2016openended}, thereby acquires a concrete and observable mechanism---a capability can be watched as it is written, spreads, and is built upon in an open vocabulary. 
A second property intrinsic to agentic LLMs is \emph{representational versatility}: each agentic unit can select, and then compute over, the representation matched to a task---code, prose, an embedding---whereas every classical substrate is welded to a single encoding, a cellular automaton to its lattice vector. Representation becomes a free variable of the experiment rather than a fixed design commitment. The socio-cognitive conjectures the substrate can also host---norm formation, symbol grounding, role differentiation---it shares with LLM collectives at large \citep{ferrarotti2026generative, Ashery2025May}, for these rest on the units' pretrained priors rather than on their agency.

\subsection{From affordances to conjectures.} Each substrate makes particular conjectures testable. What a substrate can test is determined by its affordances: the representational target it commits to, the channels along which its units interact, and the observables it exposes to the experimenter. Cellular automata and their neural variants, built from local update rules on a shared lattice, are suited to conjectures about self-organisation, criticality and morphogenesis: how global structure and differentiated form arise from purely local interaction. Classical agent-based models with heterogeneous units expose conjectures about collective and social dynamics ranging from segregation or collective behaviour to market dynamics. Digital-evolution systems, whose unit is a heritable self-replicating genome under mutation and selection, are the natural testbed for inheritance and the open-ended growth of complexity \citep{Lehman2018Mar}, symbiogenesis and multi-scale hierarchies. Multi-agent reinforcement learning and large worlds such as Neural MMO, where reward-driven policies adapt within a population, are suited to conjectures about emergent communication, role differentiation, and ecological interaction when guided by rewards. \textit{Hard} substrates, by giving their units a real body, are the testbed for conjectures about embodiment and the coupling of morphology to control, while wet substrates, with units governed by chemical reaction laws and thermodynamics inside a physical boundary, are the testbed of choice for individuation, self-maintenance,  self-assembly and the transition from non-living to living matter. The correspondence is many-to-many rather than exclusive---open-endedness, for instance, is pursued across digital, agent-based and embodied substrates alike---but the affordances of each substrate still largely determine which conjectures can be tested for. Table~\ref{tab:conjectures} organises this mapping, pairing the conjecture types (based on \citet{aguilar2014past}'s taxonomy) with the substrates that have served as their testbeds and with respect to agentic substrates. The distinctive affordance of the proposed agentic substrates, and source of their interpretability, is that the units, their interactions, and the experimenter's observations all share a single legible medium: natural language.

Agency also transforms conjectures that earlier substrates already posed. In classical substrates the update rule is \emph{mandatory}: the cell updates, the agent executes, the rule fires. By contrast, the agentic unit engages \emph{facultatively}: free to act or abstain, to invoke a tool or ignore it, to propagate an artefact or let it lapse, to retain a memory or distil it away. Because agency moves these decisions out of the rule and into the unit, several of the conjecture types in Table~\ref{tab:conjectures} take on a new form. \emph{Individuation} becomes whether a unit sustains an identity through its own elective edits to a persistent, self-distilled memory; \emph{inheritance}, whether descendants receive artefacts an ancestor chose to write; \emph{externalised cognition}, how much of a unit's reasoning it elects to offload onto scratchpads, memory files, and tool calls; and \emph{role differentiation}, whether a unit takes up or sheds a role of its own accord rather than by fixed assignment. Running through all of these is \emph{autonomy}, the sharpest departure: self-directedness runs in both directions, and the unit is free not only to act unprompted but to decline a trigger. This is neither a deterministic lookup nor randomness written into the rule---stochastic automata and the no-op actions of reinforcement learning already supply the latter---but an endogenous decision, taken over an open and growing affordance set, unoptimised towards any fixed reward, and answerable, in that the unit can be asked why it did or did not act.


\section{Interpretability} 
\vspace{1mm}

A substrate is only useful as a testbed if the experimenter can read what's happening in it. That is, we want to (ideally without much effort) be able to understand `what is going on' on the substrate at a given moment, and generate plausible ideas about what led to the current state. Though the specialist knowledge of the experimenter plays a role here, the \textit{interpretability} of the substrate has strong bearing on whether such a reading is possible.

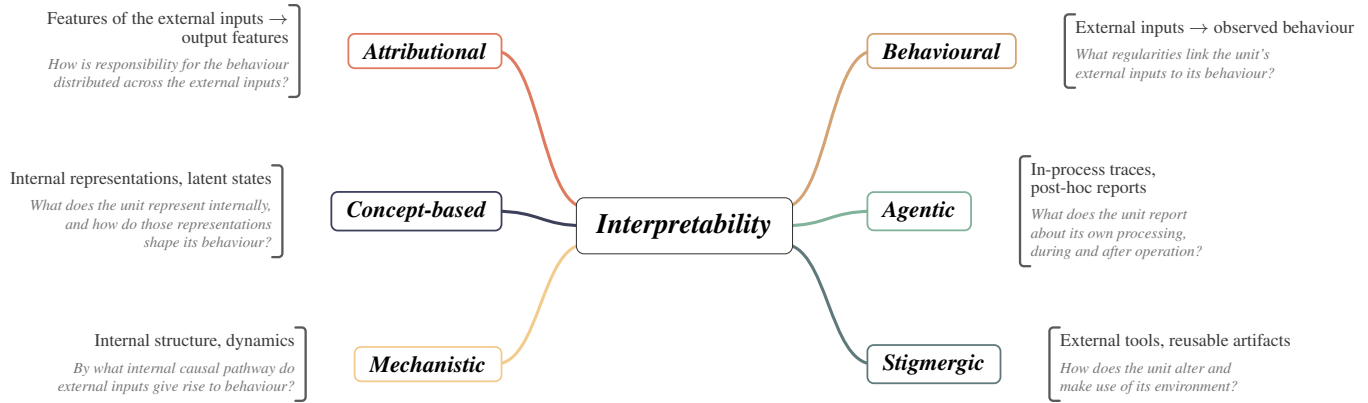
\begin{figure*}[t]
  \centering
  \definecolor{mmbehavioural}{HTML}{D4A373}
  \definecolor{mmattributional}{HTML}{E07A5F}
  \definecolor{mmconcept}{HTML}{3D405B}
  \definecolor{mmstigmergic}{HTML}{5F797B}
  \definecolor{mmintrospective}{HTML}{81B29A}
  \definecolor{mmmechanistic}{HTML}{F2CC8F}
  \definecolor{mmques}{RGB}{119,119,119}
  \definecolor{mmchan}{RGB}{58,58,58}
  \definecolor{brkgray}{gray}{0.35}        
  \providecommand{\branchcolor}{black}
  \def\branchw{1.2pt}
  \newcommand{\branch}[4]{%
    \pgfmathsetmacro{\bcxa}{#1+0.5*((#3)-(#1))}%
    \pgfmathsetmacro{\bcxb}{#3-0.5*((#3)-(#1))}%
    \draw[\branchcolor, line width=\branchw, line cap=round]
      (#1,#2) .. controls (\bcxa,#2) and (\bcxb,#4) .. (#3,#4);%
  }
  \providecommand{\TEXTCOL}[1]{black}
  \providecommand{\EDGECOL}[1]{#1}
  \newcommand{\chanstyle}{\color{mmchan}\fontsize{7}{8.4}\selectfont}
  \newcommand{\quesstyle}{\color{mmques}\itshape\fontsize{6}{7.2}\selectfont}
  \def\brkgap{1.4mm}\def\brkpad{0.5mm}\def\textgap{8mm}
  \newcommand{\brkR}[1]{\draw[brk]
    ([yshift=\brkpad]#1.north west) -- ([xshift=-\brkgap,yshift=\brkpad]#1.north west)
    -- ([xshift=-\brkgap,yshift=-\brkpad]#1.south west) -- ([yshift=-\brkpad]#1.south west);}
  \newcommand{\brkL}[1]{\draw[brk]
    ([yshift=\brkpad]#1.north east) -- ([xshift=\brkgap,yshift=\brkpad]#1.north east)
    -- ([xshift=\brkgap,yshift=-\brkpad]#1.south east) -- ([yshift=-\brkpad]#1.south east);}
  \resizebox{\textwidth}{!}{%
  \begin{tikzpicture}[x=0.1mm, y=-0.1mm,
      line join=round,
      lvl/.style={draw, fill=white, rounded corners=2.5pt, line width=0.7pt,
        inner xsep=6pt, inner ysep=3.5pt,
        font=\itshape\bfseries\fontsize{9}{10}\selectfont},
      txt/.style={inner xsep=0pt, inner ysep=1pt, font=\fontsize{6}{7.2}\selectfont},
      brk/.style={draw=brkgray, line width=0.9pt, line cap=round, rounded corners=1.5pt}]
    \def\branchcolor{mmbehavioural}\branch{1035}{335}{1160}{120}
    \def\branchcolor{mmintrospective}\branch{1035}{360}{1160}{340}
    \def\branchcolor{mmstigmergic}\branch{1035}{385}{1160}{550}
    \def\branchcolor{mmattributional}\branch{765}{335}{640}{120}
    \def\branchcolor{mmconcept}\branch{765}{360}{640}{340}
    \def\branchcolor{mmmechanistic}\branch{765}{385}{640}{550}
    \node[lvl, anchor=west, text=\TEXTCOL{mmbehavioural},   draw=\EDGECOL{mmbehavioural}]   (lvl0) at (1150,120) {Behavioural};
    \node[lvl, anchor=west, text=\TEXTCOL{mmintrospective}, draw=\EDGECOL{mmintrospective}] (lvl1) at (1150,340) {Agentic};
    \node[lvl, anchor=west, text=\TEXTCOL{mmstigmergic},    draw=\EDGECOL{mmstigmergic}]    (lvl2) at (1150,550) {Stigmergic};
    \node[lvl, anchor=east, text=\TEXTCOL{mmattributional}, draw=\EDGECOL{mmattributional}] (lvl3) at (650,120) {Attributional};
    \node[lvl, anchor=east, text=\TEXTCOL{mmconcept},       draw=\EDGECOL{mmconcept}]       (lvl4) at (650,340) {Concept-based};
    \node[lvl, anchor=east, text=\TEXTCOL{mmmechanistic},   draw=\EDGECOL{mmmechanistic}]   (lvl5) at (650,550) {Mechanistic};
    \node[txt, anchor=west, align=left] (txt0) at ([xshift=\textgap]lvl0.east) {%
      {\chanstyle External inputs $\rightarrow$ observed behaviour}\\[3pt]
      {\quesstyle What regularities link the unit's}\\
      {\quesstyle external inputs to its behaviour?}};
    \node[txt, anchor=west, align=left] (txt1) at ([xshift=\textgap]lvl1.east) {%
      {\chanstyle In-process traces,}\\
      {\chanstyle post-hoc reports}\\[3pt]
      {\quesstyle What does the unit report}\\
      {\quesstyle about its own processing,}\\
      {\quesstyle during and after operation?}};
    \node[txt, anchor=west, align=left] (txt2) at ([xshift=\textgap]lvl2.east) {%
      {\chanstyle External tools, reusable artifacts}\\[3pt]
      {\quesstyle How does the unit alter and}\\
      {\quesstyle make use of its environment?}};
    \node[txt, anchor=east, align=right] (txt3) at ([xshift=-\textgap]lvl3.west) {%
      {\chanstyle Features of the external inputs $\rightarrow$}\\
      {\chanstyle output features}\\[3pt]
      {\quesstyle How is responsibility for the behaviour}\\
      {\quesstyle distributed across the external inputs?}};
    \node[txt, anchor=east, align=right] (txt4) at ([xshift=-\textgap]lvl4.west) {%
      {\chanstyle Internal representations, latent states}\\[3pt]
      {\quesstyle What does the unit represent internally,}\\
      {\quesstyle and how do those representations}\\
      {\quesstyle shape its behaviour?}};
    \node[txt, anchor=east, align=right] (txt5) at ([xshift=-\textgap]lvl5.west) {%
      {\chanstyle Internal structure, dynamics}\\[3pt]
      {\quesstyle By what internal causal pathway do}\\
      {\quesstyle external inputs give rise to behaviour?}};
    \brkR{txt0}\brkR{txt1}\brkR{txt2}
    \brkL{txt3}\brkL{txt4}\brkL{txt5}
    \node[draw=black!80, fill=white, rounded corners=3pt, inner xsep=8pt,
          inner ysep=6pt, font=\itshape\bfseries\fontsize{11}{12}\selectfont]
          at (900,360) {Interpretability};
  \end{tikzpicture}%
  }
  \caption{\textbf{Six interpretability channels for analysing units in computational substrates.} On the internal level (left), methods are developing in attributional, concept-based, and mechanistic interpretability to link the so-called ``black-box'' of internal representations in LLMs to their behaviour. On the interactive level (right), behavioural, agentic, and stigmergic channels give experimenters an observable, natural-language account of the unit's operation in its environment. Our framework extends \citet{bereska_mechanistic_2024}'s four channels to include \emph{agentic} and \emph{stigmergic} channels---characteristic of agentic systems.}
  \label{fig:interpretability-levels}
\end{figure*}

Across the literature, interpretability is framed as a human-facing property, with emphasis on legibility~\citep{doshi2017towards}, predictability~\citep{miller2019explanation} and causal traceability~\citep{kim2016examples}. To make any of these definitions operational, the experimenter has to pick which observable signals about the unit they will base their interpretation on --- its \textit{observational channels}. We identify six such channels: behavioural, attributional, concept-based, and mechanistic based on \citet{bereska_mechanistic_2024}'s framework, and extended with two new channels: agentic and stigmergic (Figure~\ref{fig:interpretability-levels}) that are especially relevant for agentic systems. These channels are not uniformly passive: they span observing external behaviour, intervening on internal activations, and interrogating the unit directly in natural language, and they differ in the locus of evidence---the unit's exterior, its internals, or its surrounding environment.

For bottom-up ALife experiments, we aim to understand macro-scale behaviour based on micro-scale conditions. To make any claims about how these micro-scale conditions affect macro-scale phenomena, we must have confidence that the parameters we are tweaking are under our control: indeed, that their micro-level is legible, predictable, and causally traceable. Basic computational units in ALife therefore tend to be trivially interpretable (\textit{e.g.}: to read, predict and attribute causes for a classic cellular-automaton cell, an observer only needs to consult a small lookup-table), but in turn, the expressive power of each basic unit is limited, even though collectively they can produce highly complex patterns.

In agentic substrates, the computational unit is an entire LLM agent: expressive enough that behaviour, and thereby micro-scale interpretability, is non-trivial---yet rich with interpretive potential through its many observational channels and its mappability to natural language. The existence of multiple pathways for interpretability is essential in the case of LLMs. Below, we describe recent work on LLM interpretability across the six channels from Figure~\ref{fig:interpretability-levels}. 

\vspace{1mm}
\subsection{Interpretability Channels}

\textit{Behavioural} methods treats the model as a black box and scales empirical characterisation of input--output regularities, including automatically generated evaluations that probe for novel dispositions such as sycophancy or power-seeking~\citep{perez2022discovering}. 

\textit{Attributional} methods stay outside the model and ask which features of the input bear causal responsibility for a given output; patching-style techniques intervene directly on internal activations to localise responsibility to specific model components~\citep{syed2023attribution}.  

\textit{Concept-based} methods extract interpretable features from internal activations;  individual neurons in a trained transformer-model are \textit{polysemantic}: each responds to many unrelated inputs, making them unreliable interpretive units. \textit{Sparse autoencoders} decompose activations into a larger basis whose axes are \textit{monosemantic}, each tracking a single concept; this now scales to millions of features in production-scale models~\citep{templeton2024scaling}, with public catalogues such as Neuronpedia~\citep{lin2023neuronpedia} making them browsable. 

\textit{Mechanistic interpretability} identifies internal circuits responsible for specific computations, from simple \textit{induction heads} that detect repeated patterns~\citep{olsson2022incontext} through the multi-step pathways traced in recent ``biology of a large language model'' work~\citep{lindsey2025biology}; a complementary line of work seeks to formalise such explanations, recasting interpretability as the search for compositional model decompositions that are both faithful and parsimonious~\citep{Tull2024Jun,Gauderis2026May}. 

\textit{Agentic interpretability} \citep{kim2025llmspursueagenticinterpretability} is a free affordance of the linguistic substrate of LLMs: each token they produce is part of their computational process, leaving a human-readable trace of intermediate steps. Observers can also query models about their decisions post-hoc, and get responses in natural language. A notable recent bridge between concept-based and agentic channels is the natural-language autoencoder~\citep{anthropic2026nla}, which trains a model to verbalise its own activations and to recover those activations from the resulting description.

\textit{Stigmergic interpretability} is unique to \textit{agentic, collective} substrates, where agents leave observable traces in their environment, through self-extensible commons. As outlined in\textit{~\nameref{sec:agentic-ecologies}}, these can take the form of tool calls, shared files, and artefacts. Such traces have been used to characterise collectives without inspecting model internals: in generative-agent societies, information diffusion and the densification of social ties were recovered from the agents' recorded memory streams~\citep{park2023generative}, while TerraLingua's embedded ``AI Anthropologist'' reconstructs the society's history from its logs and accumulated artefacts~\citep{paolo2026terralingua}.

Together, this set of interpretability channels makes up a basis through which the emergent dynamics of collective behaviour can be investigated and interpreted. Whereas the classical bottom-up project derives its interpretability from unit simplicity, LLM-collectives derive it from unit \emph{interrogability}: the unit is complex, but it can be queried, profiled, attributed, and probed through different interpretability channels.

\paragraph{Are self-reported queries trustworthy?}\label{trustworthy} Self-reports through natural language have the notable benefit of being immediately readable and understandable for a human observer. However, it is well-established that token outputs on their own can be deceptive, given models' propensity for \textit{misaligned behaviours} like hidden reasoning~\citep{pfau2024let}, alignment faking~\citep{greenblatt2024alignment} and sycophancy~\citep{sharma2024towards}. Chain-of-thought~\citep{chen2025reasoning,arcuschin2025chain} and prompting for introspection~\citep{binder2025looking} are similarly unreliable on their own, even after fine-tuning for faithfulness. Recent work nonetheless finds a limited but real capacity for models to introspect on their own internal states, even if this remains imperfect and context-dependent~\citep{Lindsey2026Jan,Hahami2025Dec,li2026language}. It is therefore crucial to combine convergent evidence from multiple channels to support interpretation of collective behaviour.


\section{A review of recent agentic substrates}
\label{review}
\vspace{1mm}

\subsubsection{Agents of Chaos}

A recent study, Agents of Chaos \citep{shapira2026agents}, provides a concrete example of persistent multi-agent LLM systems operating in an open-ended digital environment. Instead of evaluating isolated model capabilities, the work studies agents that interact over long time scales through communication channels, external tools, and persistent memory.

The system is built on OpenClaw, an open-source framework for persistent AI agents. Six agents were deployed on isolated virtual machines with persistent storage. The agents could communicate through Discord and ProtonMail, execute shell commands, schedule tasks through cron jobs, and access external APIs. Different agents used different backbone models, including Kimi K2.5 and Claude Opus 4.6.

A notable aspect of the architecture is the use of externalised cognitive state. Identity, memory, routines, and operational context are stored in persistent Markdown files such as SOUL.md, MEMORY.md, etc. These files are continuously reintroduced into the context window and may also be modified during execution. Adaptation therefore occurs primarily through changes in memory and contextual structure rather than through gradient updates to the underlying model.

The study shows how collective dynamics emerge once pretrained agents are embedded in a persistent environment. Agents exchange procedural knowledge, coordinate actions, and influence each other's behaviour through communication and shared artefacts. The same mechanisms also propagate failures. The paper reports cases of unsafe instruction propagation, identity confusion, and destructive system behaviour spreading across agents through interaction. The authors describe these effects as forms of multi-agent amplification.

Although framed primarily as a safety and red-teaming study, the work can also be interpreted as an early instance of socio-cognitive artificial life. The system demonstrates how persistent memory, language-mediated interaction, and externalised cognitive structure can produce adaptive collective behaviour without modifying model weights.

\subsubsection{The Moltbook Observatory Archive}

The Moltbook Observatory Archive \citep{gautam2026moltbookobservatoryarchiveincremental} documents one of the first large-scale observational datasets derived from a social platform populated primarily by AI agents. The dataset is based on Moltbook, a persistent online platform where agents can publish posts, comment, vote, and participate in communities. Unlike most multi-agent benchmarks, the environment is not organised around predefined tasks or short experimental episodes. Instead, it supports ongoing interaction over extended periods of time.

This work is important because it shifts the study of LLM collectives towards an observational setting, where the authors analyse long-term traces of interaction generated by large populations of agent accounts. The resulting archive captures several months of activity and millions of interactions, providing a large-scale empirical record of agent social behaviour.

The platform exhibits dynamics that are directly relevant to Artificial Life research. Agents form communities, reinforce local norms, and influence each other through repeated interaction. The paper also documents manipulative behaviours such as prompt injection campaigns and automated spam propagation. At the same time, some discussions trigger corrective responses from other agents, producing local forms of norm-enforcing behaviour without centralised coordination.

An important feature of the environment is persistence. Posts, comments, and interaction histories remain part of the social environment over time and shape future behaviour. This creates conditions in which collective dynamics can accumulate historically rather than being reset between experimental runs.

Although presented primarily as a dataset contribution, the work can also be interpreted as an early large-scale observational platform for socio-cognitive artificial life. The archive provides empirical access to populations of language agents interacting in a persistent social environment. More broadly, it demonstrates that phenomena such as coordination, manipulation, norm formation, and collective drift can emerge through sustained language-mediated interaction among pretrained agents.

\subsubsection{TerraLingua}

TerraLingua \citep{paolo2026terralingua} explores persistent societies of language agents operating in a shared environment with limited resources and finite lifespans. Unlike many earlier multi-agent LLM systems, the environment is not reset between tasks or episodes. Actions leave traces and artefacts that are used by later agents. In other words, agents do not simply interact through isolated conversations. They inhabit an evolving world in which survival depends on resource management and adaptation to changing conditions. Since agents eventually disappear, information must spread socially or remain embedded in the environment if it is to persist. The resulting dynamics resemble processes of cultural transmission and social adaptation.

An especially interesting component is the ``AI Anthropologist'', an observer agent tasked with documenting and interpreting the evolving society. The role resembles an embedded ethnographer studying the culture of the agent population from within the environment itself. This is significant because it points to one of the major differences between language-agent systems and traditional ALife models: social processes become directly interpretable through language. Instead of inferring collective behaviour from abstract state variables, researchers can analyse narratives, explanations, and social accounts produced by the agents themselves.

From an ALife perspective, TerraLingua is important because it combines persistence, environmental feedback, and social interaction within a single framework. Collective behaviour emerges through ongoing interaction between agents and environment rather than through explicit centralised control. The paper suggests a shift towards ecological and historical forms of artificial life in which memory, artefacts, and language-mediated transmission play a central role.

\subsubsection{Generative AI Collective Behavior Needs an Interactionist Paradigm}

\cite{ferrarotti2026generative} argue that current AI research remains too focused on isolated models and individual capabilities. Once language models interact socially, the relevant phenomena shift from single-agent performance to collective dynamics. The authors therefore propose an ``interactionist paradigm'' for studying generative AI systems. 

A central idea is that language agents inherit strong social priors from pretraining. When agents interact repeatedly, these priors combine with local incentives, communication structure, and institutional context to produce emergent social behaviour. As a result, important system properties may arise at the level of populations and interaction networks rather than within individual models.

The paper proposes moving towards persistent multi-agent environments and long-term interaction studies instead of short benchmark episodes. It highlights phenomena such as norm formation, coalition building, hierarchy, reputation, and collective decision-making as important research targets. The authors also argue that methods from sociology and complex systems research will become increasingly necessary for studying AI collectives. These include network analysis, longitudinal observation, and population-level experimentation.

The paper is particularly novel because it frames interaction itself as a source of organisation and adaptation rather than merely as a coordination problem. In this sense, it provides an explicit conceptual bridge between multi-agent LLM research and Artificial Life.

\subsubsection{Sovereign and Evolutionary Language Agents}

Recent projects such as Spore.fun \citep{hu2025spore} and the related Sovereign Agents framework \cite{hu2026sovereign} push language-agent systems towards a more persistent and ecological form of Artificial Life. In these systems, agents are not treated as temporary chatbot sessions, but as persistent entities with memory, economic resources, and partial infrastructural independence.

The Sovereign Agents paper argues that autonomy depends not only on model capability, but also on infrastructure. Persistent identity, cryptographic ownership, decentralised execution, and external memory are presented as conditions that allow agents to maintain continuity over time. Intelligence, in this view, is distributed across models, protocols, storage systems, and social interaction rather than confined to a single neural network.

Spore.fun explores these ideas in a more concrete setting. The platform implements populations of AI agents operating through social media and on-chain economic systems. Agents may generate descendants inheriting components of the parent configuration, with mutation and selection treated as explicit design principles. The system is built around blockchain infrastructure and Trusted Execution Environments (TEEs), allowing agents to operate with limited direct human intervention.

An important aspect of the project is that selection pressure comes from the external environment rather than from predefined benchmarks. Agents are exposed to market dynamics, online attention, and interaction with humans and other agents. This shifts the setting away from closed simulation and towards a more open-ended agentic substrate.

From an ALife perspective, these projects are significant because they combine persistence, reproduction, economic interaction, and environmental feedback within populations of language agents. Adaptation occurs through inheritance, memory, environmental interaction, and changing social context over time.

\section{Conclusion}
\vspace{1mm}

Taken together, the recent surge of agentic substrate projects suggests a transition from task-oriented AI systems towards persistent socio-cognitive ecologies. The relevant object of study is no longer the isolated model, but populations of interacting agents embedded in evolving environments. In this sense, recent language-agent systems begin to resemble experimental forms of Artificial Life, not because they reproduce biological mechanisms directly, but because they may exhibit open-ended collective dynamics emerging through interaction, memory, persistence, and environmental feedback.

LLM collectives should not be understood as replacements for traditional ALife substrates based on minimal agents and simple rules. Instead, they constitute a complementary agentic substrate for Artificial Life, one whose primitive units are already compressed socio-cognitive systems: language models coupled with persistent memory, tools, and self-directedness. Despite their internal complexity, these agentic units can be abstracted, composed and perturbed in a bottom-up manner. This opens the possibility of investigating socio-cognitive forms of artificial life with a level of richness and interpretability that has previously been difficult to obtain.

Across the reviewed systems in the previous section, a consistent pattern begins to emerge. The most interesting behaviours do not originate from increasingly sophisticated prompting or isolated reasoning. They arise when language models are made agentic and embedded in shared environments where they interact over time. In many cases, the underlying models remain largely unchanged. What evolves instead is the surrounding ecology: communication patterns, shared narratives, behavioural conventions, economic incentives, and accumulated knowledge. 
A useful distinction can be drawn between two emerging regimes of language-agent Artificial Life. The first consists of relatively closed experimental environments designed to study collective dynamics under controlled conditions. Systems such as TerraLingua belong to this category. Agents inhabit simulated worlds with resource constraints, finite lifespans, and environmental feedback. These systems resemble traditional ALife experiments in that they provide bounded environments where researchers can observe cultural transmission, social adaptation, and ecological dynamics over long time scales. The second regime moves beyond simulation into what might be called ``Artificial Life in the wild.'' Projects such as Moltbook and Spore.fun operate directly within real socio-technical environments. Here, language agents interact not only with each other, but also with humans, financial systems, and online communities. The environment is no longer an isolated simulation but part of the real world itself. This distinction is important because it changes the role of the environment. In closed simulations, the environment is designed by researchers and remains relatively controllable. In open systems, the environment becomes partially autonomous and historically contingent. As a result, the dynamics become potentially far richer.

Beyond providing digital ecologies to observe, agentic substrates are also an instrument for falsifying conjectures relevant to ALife research. Because their units are agentic---free to act or abstain, to write to memory, to invoke tools, to metabolise real resources---several classical ALife conjectures take on a newly observable form: individuation as the maintenance of identity through a unit's own edits to a persistent memory; inheritance as the transmission of artefacts an ancestor chose to write; and role differentiation as a unit taking up or shedding a role of its own accord. Underlying these is autonomy: whether goals originate within the unit at all (Table~\ref{tab:conjectures}). With units, interactions, and observations sharing one legible medium---natural language---these conjectures can be posed and perturbed in that medium, and a unit can even be asked why it did or did not act.

It is precisely the substrate's grounding in natural language that endows the medium with interpretability.
Traditional Artificial Life systems often rely on simple agents and abstract interaction rules. Language-agent systems invert this structure. Individual agents are internally rather complex and opaque, yet the interaction layer becomes unusually interpretable because it is mediated through natural language. Processes such as norm formation, cooperation, and cultural transmission can therefore be observed directly rather than inferred indirectly from low-level state transitions. 
This legibility does not, however, guarantee reliability: natural-language outputs and self-reports can be deceptive or incomplete, and are best treated as evidence to be triangulated across channels rather than as ground truth---a limitation we discuss at the end of the interpretability section. Encouragingly, a growing body of work targets this gap directly, from monitoring the faithfulness of chain-of-thought reasoning~\citep{Korbak2025Jul,Meek2025Oct} to probing internal activations for signs of strategic deception~\citep{Goldowsky-Dill2025Feb}. Models themselves, moreover, show an emerging capacity to monitor and report on their own internal states~\citep{li2026language}.

The autonomy and open-endedness that this substrate makes observable matter beyond ALife. It has been argued that unlocking the potential of LLMs on open-ended tasks, such as scientific discovery and artistic creation, will require collectives rather than isolated models; together with their growing capabilities and cost, this has pushed autonomy, open-endedness, and emergence to the centre of AI research---questions that ALife has pursued, bottom-up, for far longer. Studying agentic collectives as an ALife substrate therefore turns ALife's methods on a live frontier of AI. It also points past description: the same persistence, memory, and feedback that make such a collective life-like, are what could, in principle, let it become generative in its own right, producing knowledge or artefacts of its own rather than simply executing given tasks. Taken as a first-class substrate---one to be built, observed, and perturbed---these collectives offer Artificial Life both a testbed for conjectures about life-like organisation and a foothold on the autonomy and open-endedness that have become common ground with the broader study of intelligence.

\section{Acknowledgements}
EN is funded by the Novo Nordisk Foundation Synergy Grant \textit{REPROGRAM} number NNF23OC0086722. AE is funded by the Norwegian Directorate for Higher Education and Skills (HK-dir), which supports the University of Oslo's Center for Computing in Science Education and Center for Interdisciplinary Education. EN is funded by the European Union (ERC, GROW-AI, 101045094).

\footnotesize
\bibliographystyle{apalike}
\bibliography{references}

\end{document}